\documentclass[letterpaper, 10 pt, conference]{ieeeconf}  % Comment this line out if you need a4paper
\usepackage[T1]{fontenc}
\usepackage{tabularx} 
\usepackage{graphicx}
\usepackage{amsmath}
\usepackage{amssymb}
\usepackage{mathtools}
\usepackage{booktabs}
\usepackage{multirow}
\usepackage{ragged2e}
\usepackage{adjustbox}
\usepackage{array}
\usepackage[hidelinks]{hyperref}
\usepackage{marvosym}  
\usepackage{url}

\IEEEoverridecommandlockouts

\title{
\large \bf
Talking Head Synthesis with Facial Landmark
Guidance via 3D Gaussian Splatting$\dagger$
}

\author{Ziheng Yang$^{1,2,3}$, Yinfeng Yu$^{1,2,3}$$^{,\mbox{\Letter}}$, and Yongming Li$^{1,2,3}$%
\thanks{\small $\dagger$This research was financially supported by the National Natural Science Foundation of China (Grant No. 62463029).}
\thanks{\small $^1$School of Computer Science and Technology, Xinjiang University, Urumqi 830017, China.}
\thanks{\small $^2$Joint Research Laboratory for Embodied Intelligence, Xinjiang University.}
\thanks{\small $^3$Joint International Research Laboratory of Silk Road Multilingual Cognitive Computing, Xinjiang University.}
\thanks{\small $^{\mbox{\Letter}}$Yinfeng Yu is the corresponding author (Email: yuyinfeng@xju.edu.cn).}%
}

\begin{document}

\maketitle
\thispagestyle{empty}
\pagestyle{empty}

%%%%%%%%%%%%%%%%%%%%%%%%%%%%%%%%%%%%%%%%%%%%%%%%%%%%%%%%%%%%%%%%%%%%%%%%%%%%%%%%
\begin{abstract}
Audio-driven digital human generation plays an important role in virtual communication, immersive interaction, and media production. With the development of Neural Radiance Fields (NeRF) and 3D Gaussian Splatting (3DGS), recent talking-head systems have obtained more faithful 3D facial geometry and appearance modeling. A remaining difficulty is that speech features mainly describe temporal acoustic patterns rather than explicit facial layouts. As a result, directly driving 3D facial deformation with audio may produce inaccurate mouth motion, weak expression details, and local artifacts. To address this issue, we propose a facial-keypoint-guided spatial enhancement module. The predicted landmarks provide structural cues for selecting and enriching spatial points around expression-sensitive facial regions. We further introduce a global landmark compensation mechanism, where the full set of keypoints is encoded into a conditioning vector to refine 3DGS attributes. This compensation supplies whole-face structural information to the underlying shape representation. Experiments under self-driven and cross-driven settings show that the proposed method improves visual quality, facial realism, and lip synchronization.
\end{abstract}

%%%%%%%%%%%%%%%%%%%%%%%%%%%%%%%%%%%%%%%%%%%%%%%%%%%%%%%%%%%%%%%%%%%%%%%%%%%%%%%%
\section{INTRODUCTION}

Speech-driven virtual facial synthesis focuses on generating realistic facial image sequences whose movements are consistent with an input speech signal. It supports a wide range of applications, including virtual avatars, film dubbing, remote communication, and real-time human--computer interaction. The problem can also be viewed as a cross-modal generation task, since acoustic information must be converted into visual motion and facial appearance. Related audio-visual studies have shown that sound cues can assist visual reasoning, spatial perception, and multimodal representation learning~\cite{YinfengICLR2022saavn,yu2023measuring,yu2025dgfnet,li2025audio}. Other multimodal works further demonstrate that external knowledge, object-level perception, and collaborative spatial reasoning are useful in vision-and-language or audio-visual navigation~\cite{yang2026beyond,yu2025dope,zhang2025iterative,yu2025dynamic,zhang2025advancing}. Robust representation learning has also been studied in missing multimodal sentiment analysis, speech synthesis vocoders, speech recognition, and speech enhancement~\cite{wang2025modality,cao2024vnet,zhang2024nonlinear,mattursun2024bss}. In addition, visual detail enhancement for segmentation highlights the value of fine-grained structural information~\cite{fu2025fsdenet}. These studies suggest that speech-driven facial synthesis requires not only cross-modal alignment, but also structure-aware and detail-preserving visual modeling.

Early talking-head synthesis methods were mainly built on 2D image generation frameworks. Such methods~\cite{DBLP:journals/corr/abs-2012-08261,das2020speech,DBLP:conf/aaai/GuZH20,DBLP:conf/cvpr/LiZWZ0CZWB023,lu2021live,DBLP:conf/mm/PrajwalMNJ20} usually learned audio-to-face mappings with generative adversarial networks (GANs). Although they can generate plausible facial textures, the absence of explicit 3D geometry makes head-pose control and spatial consistency difficult. To obtain stronger 3D awareness, later approaches~\cite{tang2025real,DBLP:conf/iccv/LiZ00023,DBLP:conf/cvpr/PengHS0ZZ00F24,DBLP:conf/eccv/LiuXWZWZ22} introduced Neural Radiance Fields (NeRF)~\cite{DBLP:conf/eccv/MildenhallSTBRN20} into talking-head generation. NeRF-based models improve view-consistent rendering and pose control, but their training and rendering costs are still relatively high, which limits their use in real-time scenarios.

3D Gaussian Splatting (3DGS)~\cite{DBLP:journals/tog/KerblKLD23} provides a more efficient representation for 3D-aware rendering. Recent 3DGS-based talking-head methods~\cite{DBLP:conf/eccv/LiZBZNZG24,DBLP:conf/mm/ChoLYHKAK24,DBLP:conf/icassp/DengZXWXSS25,DBLP:conf/icassp/FengZ0JM25} achieve high-fidelity synthesis while reducing training and rendering costs. However, the driving signal in this task is still speech, whose representation is intrinsically temporal and spatially implicit. It does not directly specify which facial regions should move or how the deformation should be distributed over 3D points. Therefore, even with a strong 3D representation, rapid mouth movements or expression changes may still lead to local artifacts and unstable facial motion.

The core challenge addressed in this work is how to guide 3D Gaussian deformation when the audio representation does not contain explicit facial spatial structure. Features extracted by ASR models such as DeepSpeech~\cite{DBLP:conf/icml/AmodeiABCCCCCCD16} and HuBERT~\cite{DBLP:journals/taslp/HsuBTLSM21} are effective for speech modeling, but they are not designed to describe facial geometry. This mismatch weakens the correspondence between acoustic patterns and facial regions, causing errors in local motion prediction. To mitigate this problem, we introduce a Spatial Enhancement Module. A pretrained audio-to-landmark model~\cite{DBLP:conf/iclr/YeJ0LHZ23} is first used to estimate 3D facial keypoints from speech. The resulting keypoint features then guide the enhancement of spatial point representations, allowing the model to focus more on regions that are closely related to expression and lip motion.

In addition to local enhancement, we further design a structure compensation mechanism based on global keypoints. Instead of using landmarks only for point-wise matching, we encode all landmarks in a frame into a global conditioning vector. This vector predicts a fine-grained compensation term for 3DGS attributes and refines the offsets produced by the backbone network. In this way, the model obtains both local keypoint-aware enhancement and global structural correction, leading to more consistent and personalized facial modeling. Experimental results demonstrate that the proposed method achieves superior performance in generation quality, facial realism, and lip-sync accuracy.

\section{Method}
\begin{figure*}[!t]
    \centering
    \includegraphics[width=0.9\linewidth]{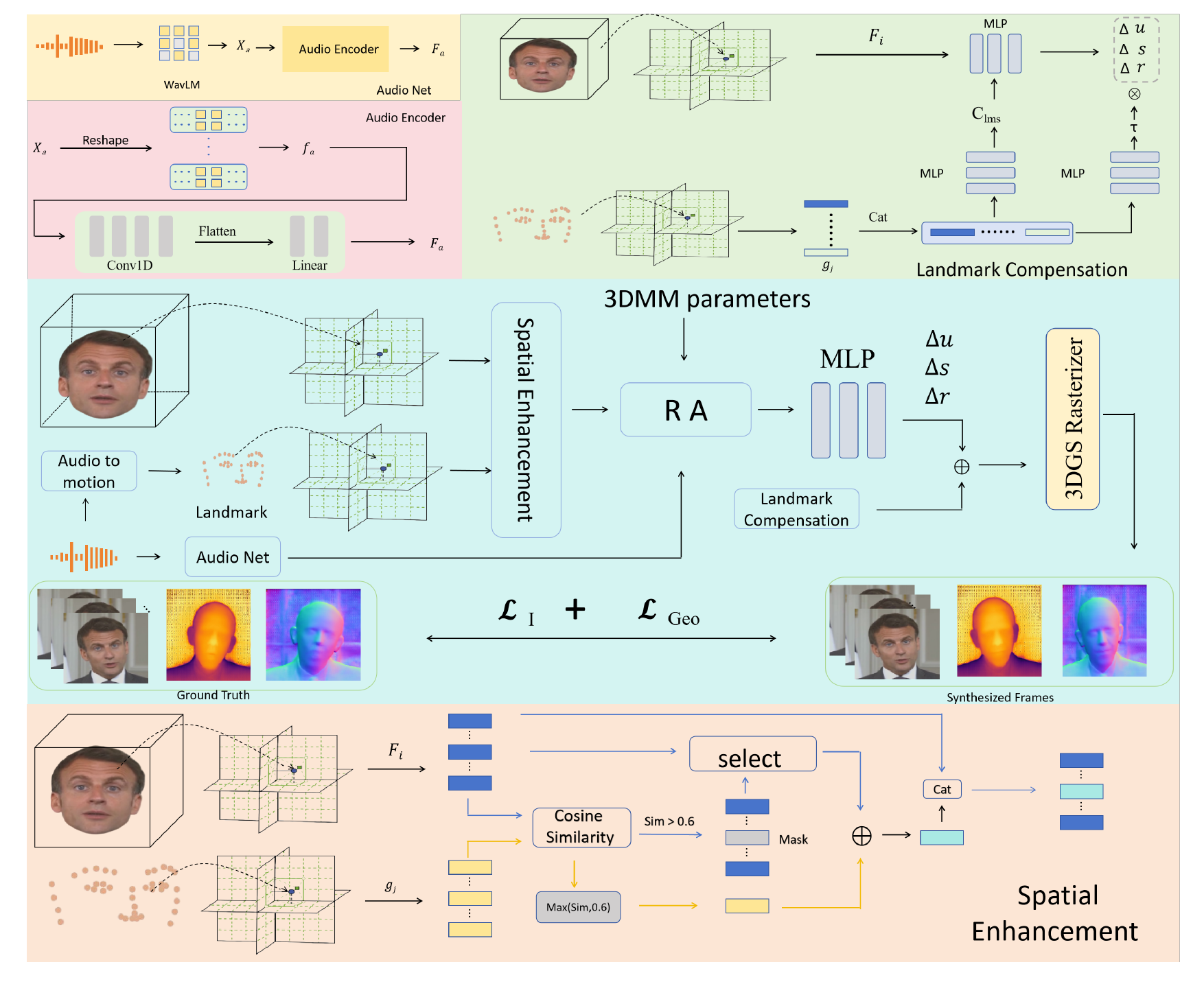}
    \caption{Overall pipeline of the proposed method. Facial keypoints predicted from audio and spatial 3D points are encoded with tri-plane features. Their similarity is used to identify spatial points that are closely related to facial structures. Region Attention (RA)~\cite{DBLP:conf/iccv/LiZ00023,DBLP:journals/pami/GuoLMH23,DBLP:conf/cvpr/HuSS18} then fuses audio, spatial features, and 3DMM parameters, including identity, shape, expression, eyelid movement, and jaw. The fused representation is used to predict 3DGS attribute offsets, while the landmark compensation branch provides global keypoint-based correction.}
    \label{fig:pipeline}
\end{figure*}

\subsection{Audio Net}

DeepSpeech~\cite{DBLP:conf/icml/AmodeiABCCCCCCD16} is widely used in audio-driven facial synthesis, but its representation capacity is limited for capturing complex speech-to-motion patterns. Moreover, it depends on large-scale supervised speech annotations. We instead use WavLM~\cite{DBLP:journals/jstsp/ChenWCWLCLKYXWZ22}, a self-supervised speech model, to extract the audio feature sequence $X_a$. Since facial motion is affected by both short phonetic changes and longer temporal context, the audio branch needs to preserve local acoustic transitions while retaining mid-range temporal information. Directly using high-dimensional WavLM features, however, increases computational cost and makes subsequent fusion less compact. Therefore, we build a lightweight Audio Encoder, as illustrated in Fig.~\ref{fig:pipeline}. A sliding window of size 2 is first applied to the WavLM features, and the windowed features are reshaped into $f_a$ to include local context. Four stacked 1D convolutional layers are then used to compress the representation. Finally, the channel dimension of the windowed feature is flattened and projected by a linear layer, producing the compact audio representation $F_a$.

\subsection{Spatial Enhancement}

Audio-driven 3D facial reconstruction is under-constrained because audio features do not provide direct spatial correspondences for facial points. When the model predicts point-wise attributes such as displacement or deformation, the absence of spatial guidance can make different facial regions semantically ambiguous. This problem becomes especially evident for rapid lip motion and expressive facial changes, where a small local misalignment may reduce the realism of the reconstructed face. Facial keypoints provide a compact description of individual facial motion. Points that are spatially or semantically close to these keypoints are more likely to belong to expression-sensitive regions, and thus should receive stronger modeling emphasis.

Based on this observation, we propose a facial keypoint-guided spatial enhancement module. The module injects landmark-based structure priors into the spatial point representation. Specifically, a pretrained audio-to-keypoint prediction network~\cite{DBLP:conf/iclr/YeJ0LHZ23} predicts 68 3D landmarks from the audio sequence for each frame. The spatial points \( \{\mathbf{x}_i\}_{i=1}^{N} \) and the predicted keypoints \( \{\mathbf{k}_j\}_{j=1}^{68} \) are then encoded by a Tri-Plane Hash Encoder \( \phi(\cdot) \). Their similarity is measured by cosine similarity:
\begin{equation}
\mathbf{f}_i = \phi(\mathbf{x}_i),\quad 
\mathbf{g}_j = \phi(\mathbf{k}_j),\quad 
S_{ij} = \frac{\mathbf{f}_i^\top \mathbf{g}_j}{\|\mathbf{f}_i\| \cdot \|\mathbf{g}_j\|}.
\label{eq:feature_mapping}
\end{equation}

A spatial point is selected for enhancement if its maximum similarity to the keypoints is greater than a threshold \( \theta \). In our implementation, \( \theta \) is set to 0.6. For each selected point, we record the most similar keypoint:
\begin{equation}
\mathcal{M} = \left\{ i \mid \max_j S_{ij} \geq \theta \right\},\quad 
j^*_i = \arg\max_j S_{ij}.
\end{equation}

The feature of the matched keypoint is added to the corresponding spatial point feature:
\begin{equation}
\mathbf{F}_i = \mathbf{f}_i + \mathbf{g}_{j^*_i}, \quad i \in \mathcal{M}.
\end{equation}

Points outside \( \mathcal{M} \) retain their original features. The resulting feature set \( \{ \mathbf{F}_i \}_{i=1}^N \) therefore keeps the original geometric distribution while assigning landmark-aware information to structurally relevant regions. This operation encourages the model to focus on the mouth, eyes, and other semantically important areas, thereby improving facial motion reconstruction and audio--lip synchronization.

\subsection{Landmark Compensation}

The spatial enhancement module provides local landmark-aware information, but its matching process is still point-wise. Local enhancement alone cannot fully describe the coordination among different facial regions, especially when the subject exhibits large expression changes or has a facial shape that differs from the training distribution. In such cases, relying only on local similarity may cause unstable or inconsistent deformation.

To introduce whole-face structural guidance, we design a keypoint-guided landmark compensation mechanism. First, the encoded features of all 68 landmarks in the current frame are concatenated and passed through a multilayer perceptron (MLP). This produces a global conditioning vector:
\begin{equation}
\mathbf{c}_{\text{lms}} = 
\psi\!\left(
\bigoplus_{j=1}^{68} \mathbf{g}_j
\right),
\end{equation}
where $\bigoplus$ denotes concatenation along the feature dimension.

The global vector is broadcast to each spatial point and concatenated with its local feature:
\begin{equation}
\mathbf{h}_i = 
\big[
\mathbf{F}_i \,\|\, \mathbf{c}_{\text{lms}}
\big],
\end{equation}
where $\mathbf{F}_i$ denotes the feature of the $i$-th spatial point.

The fused feature $\mathbf{h}_i$ is sent to two independent MLPs. The first MLP predicts fine-grained offsets for the 3D Gaussian attributes, including translation, rotation, and scale. Since directly applying an unconstrained compensation may introduce unstable motion, while a fixed global coefficient may be too weak for some facial regions, we use another MLP to predict a point-specific scaling factor $\tau_i$. This factor adaptively controls the compensation strength. The final offset is computed as:
\begin{equation}
[\hat{\mathbf{d}}_x^i,\hat{\mathbf{d}}_r^i,\hat{\mathbf{d}}_s^i]
= \text{MLP}_{\text{offset}}(\mathbf{h}_i)\cdot \tau_i,
\quad
\tau_i = \text{MLP}_{\tau}(\mathbf{h}_i).
\end{equation}

The predicted offsets are added to the backbone output as a correction term. By combining local point features with a global landmark condition, the compensation branch provides an explicit structural prior and helps produce more stable and coherent facial motion.

\begin{figure*}[!t]
    \centering
    \includegraphics[width=0.9\textwidth,keepaspectratio,height=0.75\textheight]{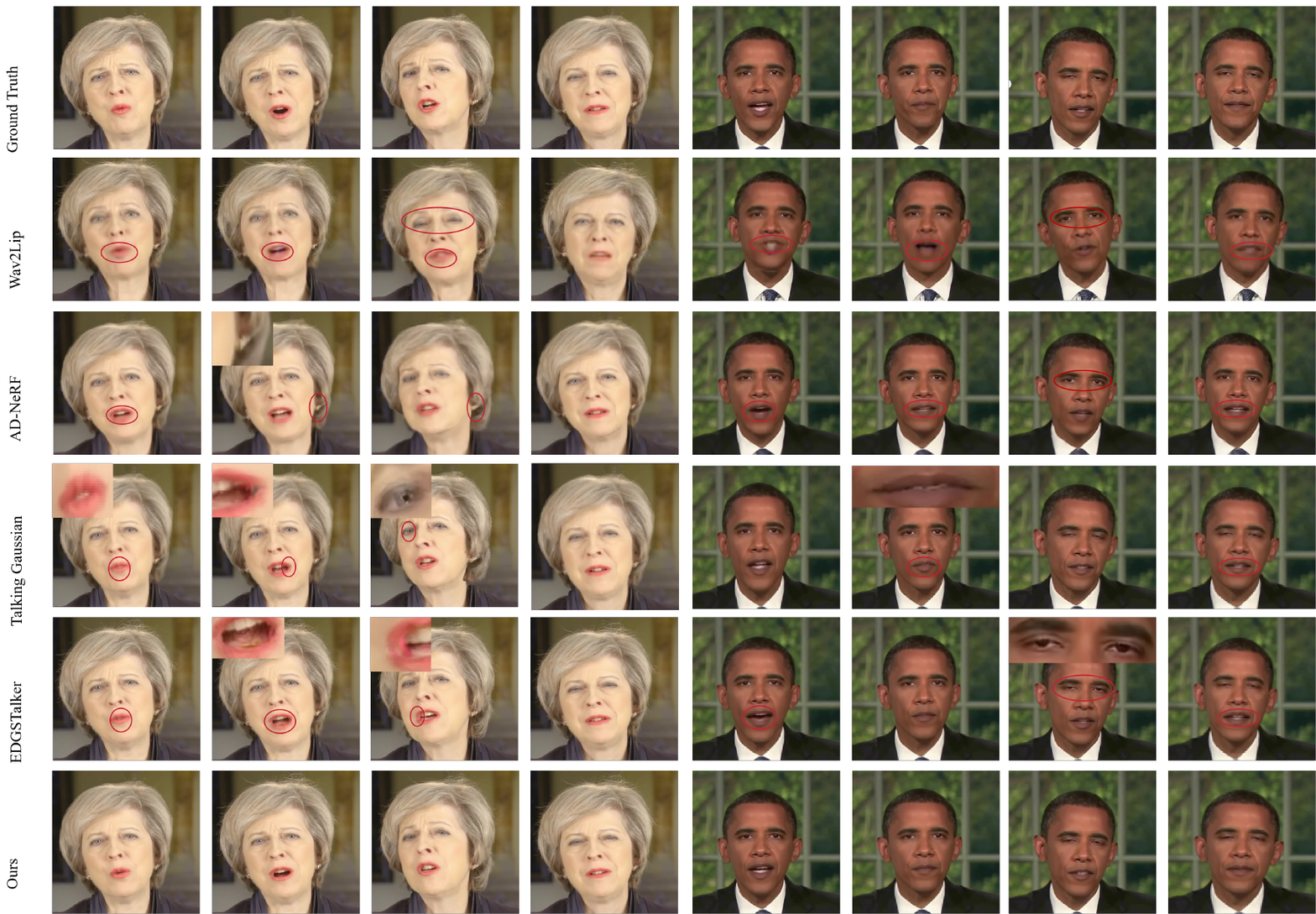}
    \caption{Qualitative comparison of talking head synthesis by different methods. Our method produces clearer facial and oral regions with fewer visible artifacts. Please zoom in for better visualization.}
    \label{fig:metric}
\end{figure*}

\subsection{Geometry Regularizer}

Sparse viewpoints and limited training frames may cause 3DGS to learn unstable geometry in unseen views. To reduce this problem, we follow the motivation of InstaGen~\cite{li2025instag} and introduce depth and surface-normal constraints. A human geometry estimator~\cite{DBLP:conf/eccv/KhirodkarBMZJSAS24} is used to estimate depth and normal maps from real images. Following~\cite{DBLP:conf/icml/ChengLYY0MWC24}, we also compute the corresponding maps from generated images and define the geometry regularization as:
\begin{equation}
\mathcal{L}_{\text{Geo}} = \lambda_D L_D(D, \tilde{D}) + \lambda_N \sum_{i=0}^{m} \sum_{j=0}^{n} \left( 1 - N_{i,j} \cdot \tilde{N}_{i,j} \right)
\end{equation}
where $L_D$ is a scale-invariant depth loss, $N_{i,j}$ denotes the normal at pixel $(i, j)$, and $(m, n)$ is the shape of image $I$. Both $\lambda_D$ and $\lambda_N$ are set to $10^{-4}$.

\subsection{Training Detail}

We follow the face--mouth decomposition architecture and optimization procedure of TalkingGaussian~\cite{DBLP:conf/eccv/LiZBZNZG24}. The training process includes static-field optimization and dynamic-field optimization. During the static stage, L1 and D-SSIM losses are used to learn a coarse head structure. During the dynamic stage, the proposed geometry-structured loss is added to improve facial geometry. The dynamic objective is:
\begin{equation}
\mathcal{L}_D 
= \text{L1}(I, I_{G})
+ \lambda\,\mathcal{L}_{D\text{-}SSIM}(I, I_{G})
+ \mathcal{L}_{Geo}
\end{equation}

\section{Experiment}
\subsection{Experimental Settings}
\subsubsection{Dataset}

Experiments are conducted on the public video datasets used in prior work~\cite{DBLP:conf/iclr/YeJ0LHZ23}. The dataset contains two male speakers, “Lieu” and “Obama”, and one female speaker, “May”. Following the common preprocessing protocol, the Obama videos are resized to \(450\times 450\), whereas the other videos are processed at \(512\times 512\).

\subsubsection{Comparison Baselines}

We compare the proposed method with two representative 2D talking-face methods, \allowbreak{}Wav2Lip~\cite{DBLP:conf/mm/PrajwalMNJ20}%
\allowbreak{} and IP\_LAP~\cite{DBLP:conf/cvpr/ZhongFCWZLL23}.%
\allowbreak{} Several 3D baselines are also included, namely AD-NeRF~\cite{DBLP:conf/iccv/GuoCLLBZ21},%
\allowbreak{} RAD-NeRF~\cite{tang2025real}, ER-NeRF~\cite{DBLP:conf/iccv/LiZ00023}, Geneface~\cite{DBLP:conf/iclr/YeJ0LHZ23},%
\allowbreak{} Talking Gaussian~\cite{DBLP:conf/eccv/LiZBZNZG24}, Gaussian Talker~\cite{DBLP:conf/mm/ChoLYHKAK24}, and DEGSTalk~\cite{DBLP:conf/icassp/DengZXWXSS25}.

\subsubsection{Implementation Details}

All experiments are performed on a Tesla T4 GPU. The full training process takes about one hour. Following the training setting of the baseline framework, the oral and facial branches are optimized for 50,000 iterations. Adam~\cite{DBLP:journals/corr/KingmaB14} and AdamW~\cite{DBLP:conf/iclr/LoshchilovH19} are used as the optimizers.

\subsection{Quantitative Evaluation}
\subsubsection{Evaluation Metrics}

We evaluate the generated results from three aspects: image reconstruction quality, facial structure accuracy, and lip synchronization. PSNR is used for reconstruction quality, LPIPS~\cite{DBLP:conf/cvpr/ZhangIESW18} measures perceptual similarity, and LMD~\cite{DBLP:conf/eccv/ChenLMDX18} evaluates landmark-level facial accuracy. For synchronization, we report SyncNet-based~\cite{DBLP:conf/accv/ChungZ16a,DBLP:conf/mm/PrajwalMNJ20} confidence score LSE-C and error distance LSE-D.

\begin{table}[htbp]
    \centering
    \caption{Quantitative comparison in the self-reconstruction setting.
    The best and second-best results are marked in \textbf{bold} and \underline{underlined}, respectively.}
    \label{tab:meteric}
    \resizebox{\columnwidth}{!}{
    \begin{tabular}{lccccc}
        \hline
        \textbf{Methods} & \textbf{PSNR}$\uparrow$ & \textbf{LPIPS}$\downarrow$ 
        & \textbf{LMD}$\downarrow$ & \textbf{LSE-D}$\downarrow$ & \textbf{LSE-C}$\uparrow$ \\
        \hline
        Ground Truth      & N/A & 0.0000 & 0.00 & 6.672 & 8.530 \\  
        \hline
        Wav2Lip\cite{DBLP:conf/mm/PrajwalMNJ20}           & 17  & 0.2954  & 4.72 & \textbf{6.609}  & \textbf{8.745} \\
        IP\_LAP\cite{DBLP:conf/cvpr/ZhongFCWZLL23}        & 16  & 0.3821  & 5.16 & 9.633            & 4.603 \\
        AD-NeRF\cite{DBLP:conf/iccv/GuoCLLBZ21}           & 26  & 0.2380  & 3.15 & 9.230            & 5.586 \\
        RAD-NeRF\cite{tang2025real}                       & 26  & 0.1250  & 2.92 & 8.902            & 5.322 \\
        ER-NeRF\cite{DBLP:conf/iccv/LiZ00023}             & 27  & 0.0864  & 2.86 & 8.712            & 6.088 \\
        Geneface\cite{DBLP:conf/iclr/YeJ0LHZ23}           & 26  & 0.1053  & 3.36 & 9.103            & 5.423 \\
        Talking Gaussian\cite{DBLP:conf/eccv/LiZBZNZG24}  & 32  & 0.0407  & 2.81 & 8.145            & 6.673 \\
        Gaussian Talker\cite{DBLP:conf/mm/ChoLYHKAK24}    & 32  & 0.0520  & 2.80 & 8.335            & 6.676 \\
        DEGSTalk\cite{DBLP:conf/icassp/DengZXWXSS25}      & \underline{37} & \underline{0.0164} & \underline{2.66} & 7.660 & 7.277 \\
        \hline
        \textbf{Ours}     & \textbf{38}  & \textbf{0.0128}  & \textbf{2.28}  & 
        \underline{7.150} & \underline{7.900} \\
        \hline
    \end{tabular}
    }
\end{table}

\subsubsection{Comparison Settings}

Two evaluation protocols are used. In the self-driven setting, each video is divided into training and testing segments. The model is trained on the training segment and driven by the testing segment from the same speaker. In the cross-driven setting, the trained model is driven by audio from other videos, which evaluates whether the method can maintain lip synchronization under audio that differs from the training utterances. Following previous practice, two audio clips from the synthesized Obama dataset~\cite{DBLP:journals/tog/SuwajanakornSK17} are used for cross-driven evaluation.

%%%%%%%%%%%%%%%%%%%%%%%%%%%%%%%%%%%%%%%%%%%%%%%%%5
\begin{table}[htbp]
\centering
\caption{Quantitative comparison in the cross-driven setting. The best result is highlighted in \textbf{bold}.}
\label{tab:cross-driven}
\begin{tabular}{l c c c c}
\hline
\multirow{2}{*}{\textbf{Methods}} & 
\multicolumn{2}{c}{\textbf{Audio A}} & 
\multicolumn{2}{c}{\textbf{Audio B}} \\
\cline{2-5}
 & \textbf{LSE-D$\downarrow$} & \textbf{LSE-C$\uparrow$} 
 & \textbf{LSE-D$\downarrow$} & \textbf{LSE-C$\uparrow$} \\
\hline
AD-NeRF          & 10.161 & 4.076 & 10.010 & 4.677 \\
Geneface         & 10.450 & 4.053 & 10.100 & 3.930 \\
Gaussian Talker  & 10.785 & 4.341 & 10.838 & 3.832 \\
Talking Gaussian & 9.969  & 4.528 & 9.722  & 4.667 \\
DEGSTalk         & 9.650  & 5.013 & 9.482  & 4.792 \\
\hline
Ours             & \textbf{9.091} & \textbf{5.736} 
                 & \textbf{8.981} & \textbf{5.506} \\
\hline
\end{tabular}
\end{table}

\subsubsection{Evaluation Results}

The quantitative results are reported in Table~\ref{tab:meteric} and Table~\ref{tab:cross-driven}. In the self-reconstruction setting, the proposed method achieves the best PSNR, LPIPS, and LMD scores, while also obtaining competitive lip-sync metrics. These results indicate that landmark guidance improves both facial structure accuracy and visual fidelity. In the cross-driven setting, our method achieves the lowest LSE-D and the highest LSE-C for both audio clips, showing stronger synchronization when the driving audio differs from the training sequence. Although Wav2Lip~\cite{DBLP:conf/mm/PrajwalMNJ20} obtains a high LSE-C in the self-driven setting, it uses SyncNet~\cite{DBLP:conf/accv/ChungZ16a,DBLP:conf/mm/PrajwalMNJ20} as an expert discriminator, which may favor this metric. Therefore, the LSE-C values of Wav2Lip should be interpreted together with other quality and structure metrics.

\begin{table}[htbp]
    \centering
    \caption{Effect of the threshold $\theta$. The best setting is highlighted in \textbf{bold}.}
    \label{tab:ablation_theta}
    \begin{tabular}{c|cccc}
        \hline
        $\theta$ & LPIPS$\downarrow$ & LMD$\downarrow$ & LSE-D$\downarrow$ & LSE-C$\uparrow$ \\
        \hline
        0.4 & 0.0147 & 2.41 & 7.203 & 7.831 \\
        0.6 & \textbf{0.0128} & \textbf{2.28} & \textbf{7.150} & \textbf{7.900} \\
        0.8 & 0.0138 & 2.35 & 7.176 & 7.856 \\
        \hline
    \end{tabular}
\end{table}

\subsection{Qualitative Evaluation}

Fig.~\ref{fig:metric} gives a visual comparison under the self-driven setting. The red boxes highlight regions with blur, artifacts, or unnatural facial details. Compared with the baseline methods, the proposed method produces clearer mouth shapes and more stable facial regions. The results also show better agreement between lip motion and speech, especially around the oral region.

\subsection{Ablation study}

Table~\ref{tab:RCFN-ablation} analyzes the contribution of each component. When the proposed Audio Net is replaced with DeepSpeech, LMD and LPIPS change only slightly, but LSE-D and LSE-C degrade more clearly. This suggests that the proposed audio representation is more helpful for aligning speech with lip motion. Removing the Audio Encoder also weakens the synchronization metrics, indicating that WavLM features require additional temporal compression before being used for facial motion prediction. Without the Spatial Enhancement (SE) module, the model obtains worse LMD and LPIPS, and the lip-sync metrics also decline slightly. This confirms that SE contributes to local spatial representation and detailed visual reconstruction. Removing the Landmark Compensation (LC) module causes a larger increase in LMD, showing that global keypoint compensation is important for landmark accuracy and motion consistency. The full model combines these components and achieves the best overall result.

\begin{figure}[t]
    \centering
    \includegraphics[width=\linewidth]{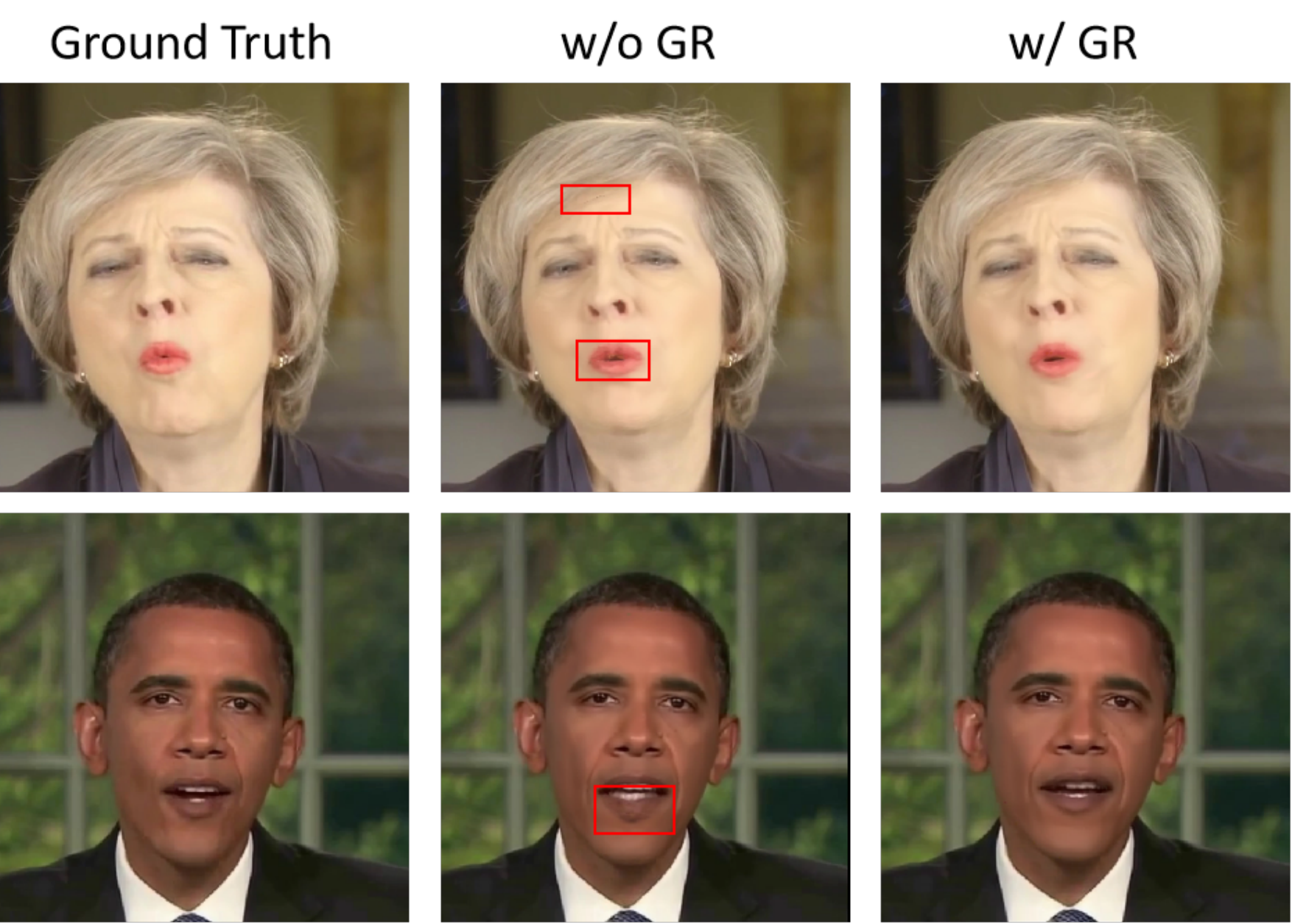}
    \caption{Ablation Visualization of Geometry Regularizer.}
    \label{fig:loss}
\end{figure}

The Geometry Regularizer mainly improves visual quality rather than producing large numerical gains. As shown in Fig.~\ref{fig:loss}, adding this regularizer reduces local facial artifacts and makes the generated faces more natural. We further evaluate different values of $\theta$ in Table~\ref{tab:ablation_theta}. With $\theta=0.4$, nearly half of the spatial points are enhanced, which reduces the selectivity of the module. With $\theta=0.8$, only a small number of points are selected, so the coverage of important facial regions becomes insufficient. The setting $\theta=0.6$ selects about 20\% of the points and covers the mouth region and its neighborhood, providing a better trade-off between selectivity and spatial coverage.

\begin{table}[htbp]
\centering
\caption{Component ablation results.
DS refers to DeepSpeech.}
\label{tab:RCFN-ablation}
\setlength{\tabcolsep}{2pt}
\renewcommand{\arraystretch}{0.85}
\begin{tabularx}{\columnwidth}{>{\bfseries}l c c c c}
\toprule
\textbf{Setting}  & \textbf{LMD}$\downarrow$ & \textbf{LPIPS}$\downarrow$ & \textbf{LSE-D}$\downarrow$ & \textbf{LSE-C}$\uparrow$ \\
\midrule
Replacing Audio Net with DS & 2.36 & 0.0135 & 7.663 & 7.443 \\
w/o Audio Encoder & 2.39 & 0.0138 & 7.502 & 7.585 \\ 
w/o SE & 2.45 & 0.0153 & 7.221 & 7.801  \\
w/o LC & 2.52 & 0.0150 & 7.256 & 7.818   \\
w/o GR & 2.28 & 0.0131 & 7.203 & 7.862  \\
\midrule
Full model & \textbf{2.28} & \textbf{0.0128} & \textbf{7.150} & \textbf{7.900}   \\
\bottomrule
\end{tabularx}
\end{table}

\section{Conclusion}

This paper introduces a facial-landmark-guided 3DGS framework for speech-driven talking-head synthesis. The spatial enhancement module uses predicted landmarks to enrich spatial point features around expression-sensitive regions, while the landmark compensation mechanism provides global structural correction for 3DGS attribute prediction. The Geometry Regularizer further improves the stability of the generated facial geometry, and the lightweight Audio Net provides compact speech representations for motion prediction. Experiments in both self-driven and cross-driven settings show that the proposed design improves visual quality, facial structure accuracy, and lip synchronization. The ablation results further verify the contribution of each component.

\bibliographystyle{IEEEbib}
\bibliography{refs,strings}

\end{document}